\documentclass[conference, letterpaper]{IEEEtran}
\usepackage[numbers,sort&compress,square]{natbib}
\usepackage{url}
\usepackage{mathtools} 
\usepackage{amssymb} 
\usepackage{amsmath}
\usepackage{tabu}
\usepackage{float}
\usepackage[font=footnotesize,labelfont=bf]{caption}
\usepackage{tablefootnote} 
\usepackage[paper=letterpaper,top=1in,bottom=0.75in,right=0.75in,left=0.75in]{geometry}
\usepackage{lipsum}
\usepackage[pdftex]{graphicx}
\begin{document}

%
\title{\huge Paradigm Shift in Continuous Signal Pattern Classification:\\
	Mobile Ride Assistance System for two-wheeled Mobility Robots}


\author{\IEEEauthorblockN{Ali Boyali,
		Naohisa Hashimoto and Osamu Matsumoto}
	\IEEEauthorblockA{National Institute of Advanced Industrial Science and Technology\\
		Tsukuba, Ibaraki, Japan\\
		E-mail:  \{ali-boyari\}\{naohisa-hashimoto\}\{matsumoto.o\}@aist.go.jp}}

\maketitle

\begin{abstract}

In this study we describe the development of a ride assistance application which can be implemented on the widespread smart phones and tablet. The ride assistance application has a signal processing and pattern classification module which yield almost 100\% recognition accuracy for real-time signal pattern classification. We introduce a novel framework to build a training dictionary with an overwhelming discriminating capacity which eliminates the need of human intervention spotting the pattern on the training samples. We verify the recognition accuracy of the proposed methodologies by providing the results of another study in which the hand posture and gestures are tracked and recognized for steering a robotic wheelchair.  
 
\end{abstract}


\section{Introduction}

Two-wheeled standing type Personal Transporters (PT), such as Segway (TM) and Toyota Winglet have been seen an alternative transportation mode in the short range excursions and gained acceptance from a broad range of users in many countries. Driven by electric motors and having a good travelling range that varies between 10-38 km depending on the vehicle type and brand, the PTs are preeminent for the future of transportation. 

Even though, the PTs have been developed for more than a decade, in many countries except USA, the use of these vehicles on the public road related areas is prohibited by legislation and traffic regulations \cite{segwaycomp}. The lingering controversy on the vehicle categorization limits the use of PTs only within the designated areas or private properties. The upper speed level of PTs and lack of braking and signaling equipment, insufficient number of reports and experimental results make the vehicle categorization impossible within the current regulations of the these countries confusing the  regulation bodies and law makers. 

On the other hand, these kind of mobility robot technologies are necessary and desirable to reduce the traffic congestions, the use of single occupancy vehicles for short distance trips and to increase the independent mobility of the people such as the elderly. The aging society have been the very fact that the many developed countries face today. The increasing population of the elderly who are 65 years and older cohort in the developed countries have tremendous impact in nation's economic vitality due to the increasing health care and pension payment and the lack of care givers that can meet the emerging demand \cite{euromon, unhr}. Aging processes do not only have impact on the economic vitality and social system, but also the elderly themselves. The decreasing motor and cognitive skills, the elderly are prone to social isolation, depression and poverty due with the less transportation options to the nearest public hubs and city amenities than the able-bodied individuals.

In this study we describe the development of an accurate and robust signal classification framework that suits for any type of sensor domain to implement a ride assistance system on the mobile smart-phones and tablets. The application on a smart phone or tablet which is attached to the handle bar of a PT continuously observe the sensor measurements and classify the breaking maneuvers real-time and notify the rider if the maneuver is dangerous while recording the ride motion characteristics. The signal pattern classification framework is not limited the breaking mode but it can be implemented for various application domains such as classification of lateral maneuvers or determining the road surface roughness as we elaborate in the proceeding sections. 

The signal processing and classification module is the core of the ride assistance system which also show the location and velocity information to the rider by producing warnings and notifications on the screen. We designed a complementary filter for sensor fusion for this purpose. The acquired measurements from gyroscope and accelerometer are filtered out to obtain reliable orientation and motion parameters for the classification module.

We used the Block-Sparse \cite{li2013robust,boyali2014block} and Collaborative Representation based Classification \cite{zhang2012collaborative,zhang2011sparse} (B\_SRC and CRC) methods for the ride assistance system and compared the performance of classification accuracy and computation times for the mobile device. The success of these classification methods are highly dependent on the reliable training dictionaries with a powerful discriminating capability. Building such kind of dictionaries for the highly noisy signals require defining the start and end point of the training signal. However this task is highly difficult due to the fact that in most cases the pattern boundaries are not distinct. One of the novelty we propose and demonstrate by this paper is building a dictionary with a high discriminating power without spotting the signal pattern and using this dictionary for a continuous recognition of the performed maneuver. The approaches eliminates signal spotting both in training and real-time phases.  
 
We organized the rest of the paper as follows. In Section II, we cogently review the mobile robot studies, then describe the robotic zone designated for investigation of the robot-society interaction. After briefly describing  the configuration of the mobile tablet and mobility robot system, we detail on how we implement a complementary filter in an easy way in Section III. The section IV is dedicated to the classification systems and building training dictionary. The experimental results and discussion is given in the section V. The paper ends with the conclusion and future works in the conclusion section.

\section{Prior Works and Project Description}

A scant number of studies in the literature investigate the interaction of the mobility robots with the other road users on the public road related areas. The limited literature and experimental data are not sufficient to guide the regulation authorities and developers to make decisions and improvements collaboratively. Among these studies, the authors in \cite{miller2008segway} report the approaching distance the mobility robots to the other road sharers at the various speed profiles with the recruited novice and expert riders. The stopping distance for the different driving maneuvers, the braking and response time of the riders are assessed in a similar study \cite{nishiuchi2013segway}. In the study \cite{hashimoto2013experimental}, the approaching distance of the other road vehicles and pedestrians to the mobility robots including Segway and robotic wheelchairs are presented. All of the experimental studies are carried out under the controlled environments with the limited resource that doesn't contribute to sufficient statistics for the real world ride characteristics. In addition to the experimental studies, the reports \cite{actgovern,canada} release the opinion of the many stakeholders and institutions as well as the subjective assessment of the recruited riders. 

\begin{figure}[h]
	\centering
	\includegraphics[width=\columnwidth, keepaspectratio=true]{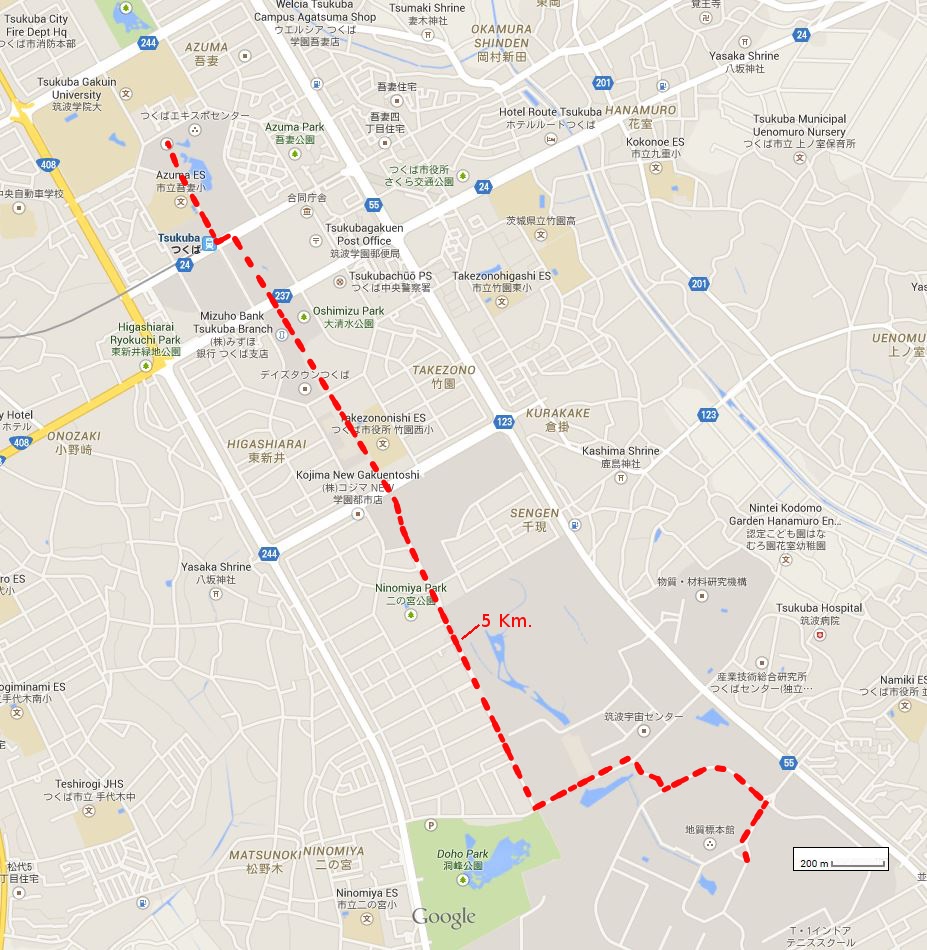} 
	\caption{Tsukuba Robotic Zone}
	\captionsetup{justification=centering}	
	\vspace{-1em}
	\label{fig:f1}
\end{figure}

The crux of our motivation is to create a ride assistance system which analyzes the ride dynamics, lateral and longitudinal maneuvers as well as the rider's behaviors and produce immediate warnings when a dangerous situations arise while continuously gathering ride characteristics, interaction of the rider with other road sharers in the designated robotic zone (Fig. \ref{fig:f1}) to form a national database on the robot-society interaction. 

\begin{figure}[h]
	\centering
	\includegraphics[width=\columnwidth, keepaspectratio=true]{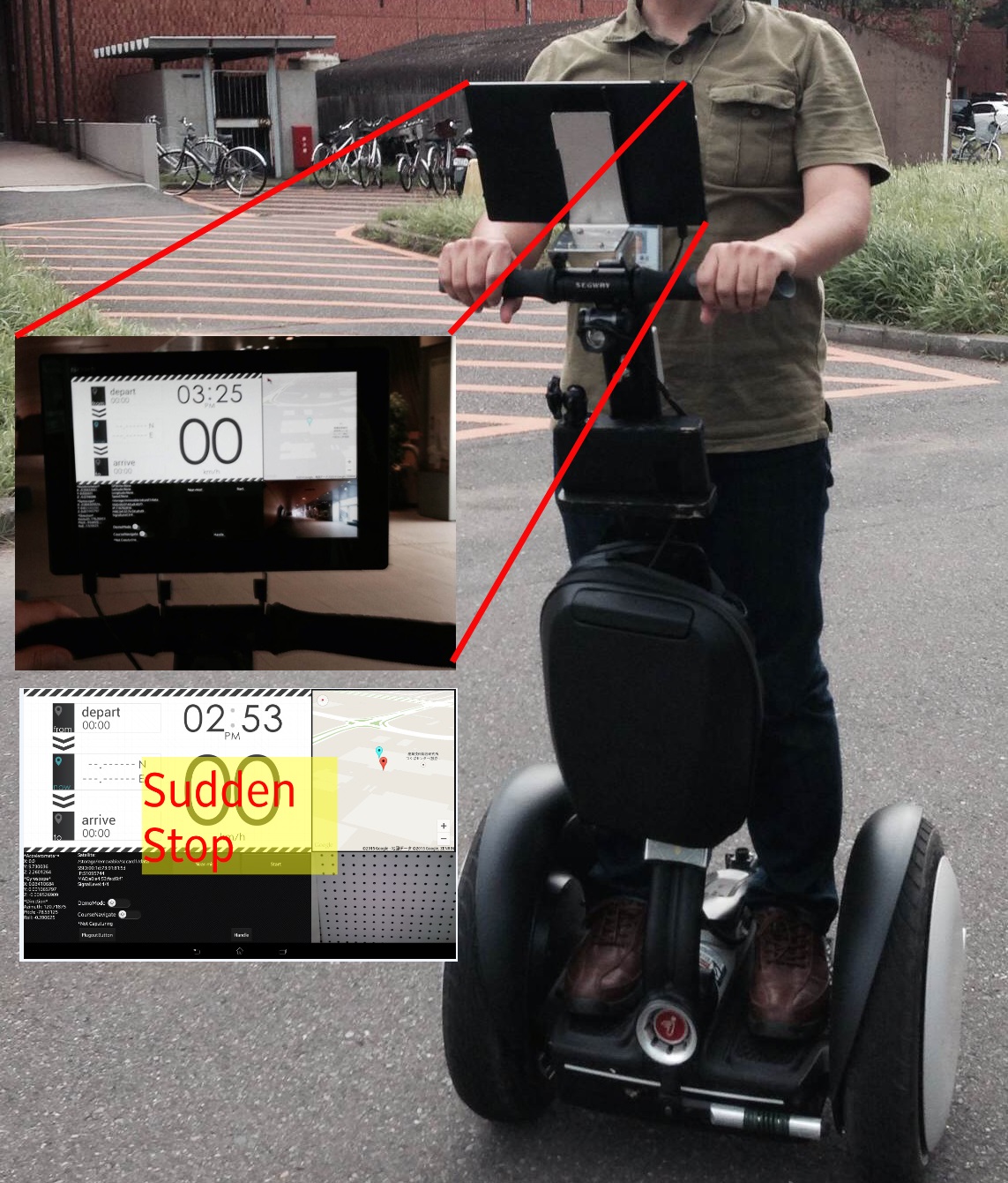} 
	\caption{Segway and Android Tablet}
	\captionsetup{justification=centering}	
	\vspace{-1em}
	\label{fig:f2}
\end{figure}

The robotic zone in Tsukuba is the only designated zone in Japan in which the interaction of the robotic mobility devices with the other road sharers is investigated by various research institutes. The robotic mobility device, Segway is used in the experiments, the results of which is given in this paper. An Android tablet the screen facing up to the riders is attached on the top of handle bar (Fig. \ref{fig:f2}) \cite{boyaliUn}.  

The tablet connected to Segway sharing system shows the time to destination, speed and current location as well as the sensor readings on the screen. The handle bar of Segway is under continuous excitations induced by the steering actions and road irregularities. The signal processing and classification module is integrated to the mobile application to align the coordinate systems of the tablet and Segway to obtain the motion related parameters.

\section{Signal Processing and Filters}

\begin{figure}[h]
	\centering
	\includegraphics[scale=0.35, keepaspectratio=true]{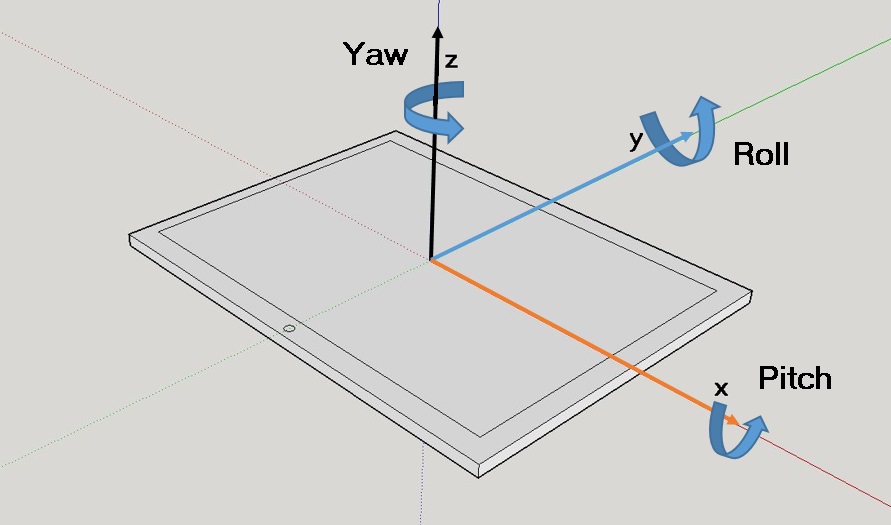} 
	\caption{Segway and Android Tablet}
	\captionsetup{justification=centering}	
	\vspace{-1em}
	\label{fig:f3}
\end{figure}

The classification methods used in the study make use of the gyroscope output pitch rate and the pitch orientation of the handle bar as the most discriminating features for the braking mode classification. The other sensor measurements such as that of the accelerometer are also used for other classification and notification purposes. A rotation matrix is required in order to align the tablet coordinate system (Fig. \ref{fig:f3}) with the motion coordinates of the robot to rotate the accelerometer measurements subtracting the gravity vector. Since, the handle bar of Segway has only two Degrees of Freedom (DOF), the pitch and roll rotations are the only input parameters for this rotation matrix.   

\begin{figure}[h]
	\centering
	\includegraphics[width=\columnwidth, keepaspectratio=true]{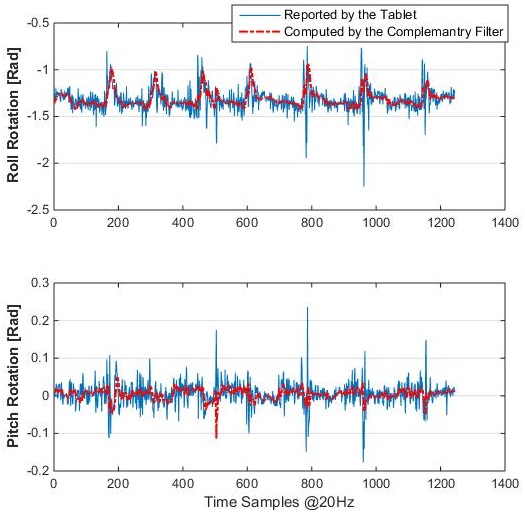} 
	\caption{Reported and Filtered Orientation Angles}
	\captionsetup{justification=centering}	
	\vspace{-1em}
	\label{fig:f4}
\end{figure}

Although, Android Software Development Kit (SDK) provide access to the orientation, position and motion sensor information, the reported acceleration and orientation values are only reliable at low frequencies or when the device rests at stand still as stated in the Android SDK documents \cite{androids}. In Fig. (\ref{fig:f4}) the reported and filtered roll and pitch orientation are compared for an experiment in which the rider does subsequent sudden breaking maneuvers where high frequencies prevail. As seen in the figure, the reported orientation angles are very noisy with respect to the measurements obtained by a complementary filter. 

If the reported orientation angles are used for the rotation matrix, the noise factor of the accelerometer measurements to be rotated are amplified by this transformation. Thus, we apply a low pass, third order Butterworth filter to reduce the noise on the acceleration values which are then reserved for other classification purposes such as road surface road surface detection. 

The most discriminating features are the pitch rate and orientation angle which is obtained by a complementary filter for the breaking mode classification addressed in this paper. A complementary filter sums the transfer functions of measurement processes which are mathematically complement to each other \cite{higgins1975comparison}.  Micro Electro-Mechanical Sensors (MEMS) integrated on the mobile computing devices have advantages and disadvantages at certain frequency intervals. Gyroscope measurements have low noise level at the higher frequencies while the accelerometer component give relatively reliable measurements at the lower frequencies. As a variant of the steady state Kalman filter, a complementary filter can be implemented by a Proportional and Integral (PI) controller \cite{yoo2011gain}.

In this application we design a complementary filter to obtain reliable roll and pitch orientation angles. These variables can be computed from either of the sensor's measurements; by integrating the gyroscope readings or using the trigonometric functions on the accelerometer as the component of the gravity vector appears on each axis of the acceleration data. These trigonometric functions to derive the pitch ($\theta$) and roll ($\phi$) angles from the measured acceleration on the each axis ($a_x, \quad a_y, \quad a_z$) are given  Eq. (\ref{eq:e1}). 

\begin{equation}
\phi = atan2(a_{y}, a_{z}),\quad  \theta=atan2(-a_{x}, \sqrt{a_{y}^2+a_{z}^2})   
\label{eq:e1}
\end{equation}

\begin{figure}[h]
	\centering
	\includegraphics[width=\columnwidth, keepaspectratio=true]{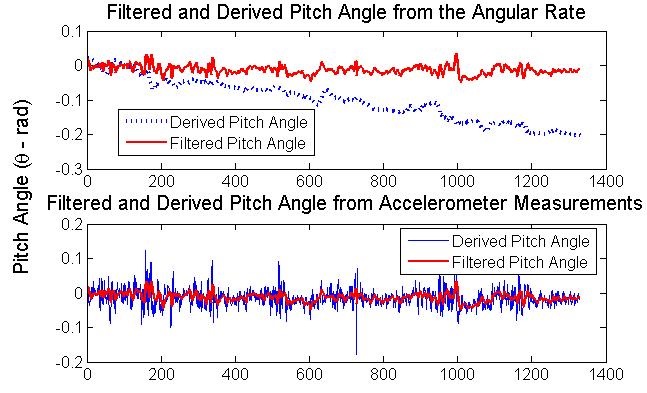} 
	\caption{Pitch Angle Derived from the Sensors and Filtered}
	\captionsetup{justification=centering}	
	\vspace{-1em}
	\label{fig:f5}
\end{figure}

The derived pitch angles from gyroscope and accelerations in this way and the output of the complementary filter are compared in Fig \ref{fig:f5}. As seen in the figure  the computed orientation angles from the gyroscope measurements exhibits drift due the integrated noise, whereas those from the accelerometer is jittery.

\begin{figure}[h]
	\centering
	\includegraphics[width=\columnwidth, keepaspectratio=true]{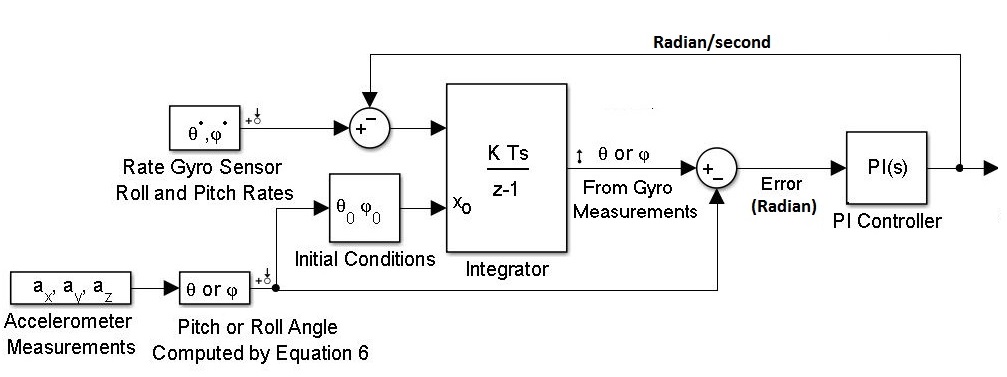} 
	\caption{Complementary Filter Simulink Block Diagram}
	\captionsetup{justification=centering}	
	\vspace{-1em}
	\label{fig:f6}
\end{figure}

Fig. \ref{fig:f6} shows the block diagram of the complementary filter Matlab Simulink model. The error between the computed orientation angles by the gyroscope and accelerometer sensor measurements are used as the input for the PI controller in the block diagram. The output is the control input for the process.  

The transfer function for the estimated pitch and roll angles becomes;

\begin{equation}
{\phi}_{f} =\frac{1}{s}\dot{{\phi}_g} +\frac{K_{p}}{s}({\Phi_a}-{\Phi_{f}}) +\frac{K_{i}}{s^2}({\Phi_a}-{\Phi_{f}})
\label{eq:2} 
\end{equation}

After a series of algebraic manipulations, the transfer function for the roll and pitch angles is obtained as in follows.
 
\begin{equation}
{\phi}_{f} =\frac{s^2}{s^2+K_{p}s+K_{i}}\big(\frac{1}{s}\dot{\phi_g}\big) +\frac{K_ps+K_i}{s^2+K_ps+K_i}\phi_a
\label{eq:3}
\end{equation} 

In the equation $\phi_f$,  $\phi_g$ and $\phi_a$ represent the filtered roll and pitch rotations obtained from the gyroscope and accelerometer measurements. The same difference equation is used to compute both orientation angles. The filter coefficients $K_{p}$ and $K_i$ must be defined. We used the PID tuning toolbox of Matlab to find the optimum filter coefficients. Once the transfer function and it's coefficients are obtained, the difference equation for the digital filtering can be computed using the bilinear transformation. The Matlab's PID tuning toolbox gives the optimum coefficients as $K_p=7.5924$ and $K_i=20.7015$. Upon applying the bilinear transformation, the difference equation for the roll rotation  is obtained as:

\begin{align*}
\label{eq:99}
\phi_f=\frac{(0.907z^{-1}-1.814z^{-2}+0.819z^{-3})\phi_g}{1-1.8091z^{-1}+0.8188z^{-2}}+\\
\frac{(0.093z^{-1}+0.0048z^{-2}-0.0882z^{-3})\phi_a}{1-1.8091z^{-1}+0.8188z^{-2}}
\end{align*}

\begin{figure}[h]
	\centering
	\includegraphics[width=\columnwidth, keepaspectratio=true]{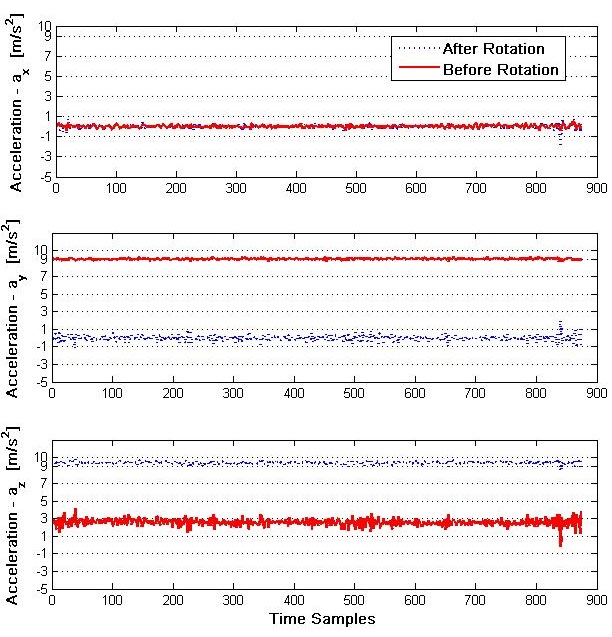}
	\caption{Aligned and Unaligned Accelerometer Measurements when the Segway at stand still}
	\label{fig:rotated_acc}	
	\vspace{-1em}
\end{figure}

The rotation matrix is obtained using these filtered pitch and roll rotations. The accelerometer measurements are multiplied with the rotation matrix to align the sensor readings to the Seqway's motion coordinate system. The rotated accelerometer measurements are given in Fig. (\ref{fig:rotated_acc}). 

The dotted curves represent  the aligned measurements in the body coordinate system whereas the continuous curves represent the unaligned measurements. The acceleration values were recorded while the Segway was in a stand still position. As seen in Fig. (\ref{fig:rotated_acc}), the gravity value measured along the $z$-axis of the Android device fluctuates around Earth's gravity value after rotation. As there is no lateral and longitudinal acceleration, the acceleration values on the $x$ and $y$ axes fluctuate around the zero line as expected.

\section{Classification Methods and Training}

\subsection{Classification Methods}
The Sparse Representation based Classification (SRC) \cite{wright2009robust} and CRC assume that any observed signal $y$ can be approximated as a linear combination of the sample signals from the same classes. The resulting system of linear equation systems ${y}=Ax$ is then solved using a combination of vector norms.  The general objective function of the optimization problem is given in Eq. (\ref{eq:e5}) where $A=[A_1,A_2,\ldots,A_n]  \in \mathbb{R}^{mxn}$ represents the training dictionary in which the training samples from every classes are put as a column vector and $\lambda$ is the Lagrange penalty coefficient.

\begin{equation}
\label{eq:e5}
\min_{x}\quad \hat{x}=\Arrowvert y-Ax\Arrowvert_{q}+\lambda\Arrowvert x \Arrowvert_p
\end{equation}

In the SRC method, the norm values $q=1\quad or \quad 2$,  and $p=1$ are used depending on the noise and outlier conditions, whereas in  the Least Square Regularization version of the CRC method (CRC\_RLS) \cite{zhang2011sparse}, the norm values $q=p=2$ are used. In this case the objective function becomes: 

\begin{equation}
\label{eq:e3}
\min_{x}\quad \hat{x}=\Arrowvert y-Ax\Arrowvert_{2}^{2}+\lambda\Arrowvert x \Arrowvert_2^{2}
\end{equation}

The label of the observed signal is obtained using the objective function given in Eq. (~\ref{eq:e6}) where $\delta_{i}:\mathbb{R}^{n}\rightarrow\mathbb{R}^{n}$ is the selection operator that selects the coefficients of $i^{th}$ class while keeping other coefficients zero in the solution vector $\hat{x}$.

\begin{equation}
	\label{eq:e6}
	\min_{i}\quad r_{i}(y)=\Arrowvert y-A\delta_{i}(\hat{x})\Arrowvert_{2}
\end{equation}

The SRC relies on an overcomplete dictionary whereas for the CRC\_RLS, the dictionary might be undercomplete, as the observed signal is approximated by using the training samples from other classes collaboratively. The solution for the objective function given in Eq. (\ref{eq:e6}) is a Ridge regression given in Eq. (\ref{eq:e7}). 

\begin{equation}
\label{eq:e7}
\hat{x}=({A^T}A+{\lambda}I)^{-1}{A^T}y
\end{equation} 
 
Both of the methods yield high classification accuracy as well as a training dictionary with a high discriminating capacity. The iterative nature of the $\ell_1$ solvers make the SRC based methods slower then the CRC method in which the pre-computed regression operator $P_{\lambda}=({A^T}A+{\lambda}I)^{-1}{A^T}$ is only computed once because it is independent of the observed signal $y$ and stored for further computations. Due to these reasons, the computation times are beyond comparison that nominates the use of CRC method as a good alternative on smart phones and tablets. We compared the computation times in our simulations and found out that the B\_SRC takes 112 seconds for complete the computations for 1000 time samples with the minimum number of the iterations while the CRC takes 0.39 seconds. 

\subsection{Training - Finding Best Representative Patterns}

\begin{figure}[h]
	\centering
	\includegraphics[width=\columnwidth, keepaspectratio=true]{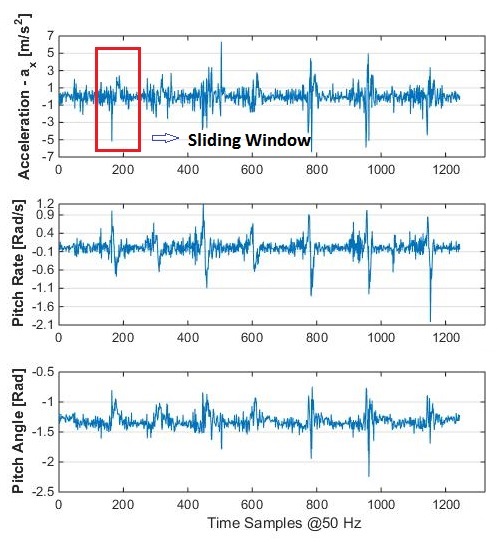}
	\caption{Aligned and Unaligned Accelerometer Measurements during a Sudden Stop Experiment}
	\label{fig:segtrain}
	\vspace{-1em}		
\end{figure}

In real time signal pattern classification applications, two important issues arise. First, the signal that is observed and sampled are acquired by a sliding window with a fixed width. This acquisition procedure stipulates putting representative signal samples for every observed pattern in the training dictionary for sake of classification accuracy. Ensued by the second, picking these representative samples on the training signal on which the boundaries of the class samples are vague. Alluded issues deserve some elaboration. Let's assume that we have a multi-dimensional signal as given in Fig.  (\ref{fig:segtrain}). 

The signals are sampled at a frequency of 20 Hz and the observed portion of a signal is acquired by a fixed length sliding window. Fig. (\ref{fig:segtrain}) shows three measurements; acceleration along the $x$-axis, pitch rate and angle collected for a sudden breaking experiment. In the experiments, the rider accelerates, reaches a constant velocity then try to stop the Segway immediately by harsh breaking maneuver seven times. When we analyze the signal we ostensibly expect and observe two breaking states which are cruising (no braking) and sudden breaking while there are three breaking patterns. The normal breaking is hidden in the sudden breaking regions on the signal and the mobility robot passes from normal breaking state for a short time before and after the sudden stop maneuver. As seen in the figure, in addition to indistinct noisy signal boundaries between the cruising and sudden breaking regions, the signal patterns belonging to either of these classes are not alike. 
 
In order to build a dictionary on the given training signal, the start and end points of the patterns must be distinguished exactly that requires a diligent and tedious effort. Albeit one built does not necessarily guarantee a high discrimination power. On the other hand,  the sliding window captures some portions from any of two states with the varying length during transition that deteriorates the classification accuracy. 

The remedy to these problems encountered in building a training dictionary for the real-time applications is to create a matrix, the columns of which is captured by a sliding window, then clustering the training matrix into the union of subspaces that generates the signal patterns in the classes. The resulting training matrix takes the form of a block Hankel matrix in which the neighboring columns are very similar to each other. In this regard, the clustered matrix columns according to their generating union of subspaces can be further clustered applying the same procedure, if any training class is a combination of any two classes such as in the case that sudden breaking patterns. When this training framework is implemented in a nested manner, all the representatives from each class are distilled into their pure class range successfully.

The subspace clustering algorithms have gained impetus during the era of Compressed Sensing (CS) research field which seeks the sparsest solutions to the linear representation problems. Similar to the CS problem setup, the form of which is given for SRC, the subspace clustering methods aim to represent the signal patterns stacked into a dictionary by the linear combination of similar signals generated by the same subspace by seeking the sparsest solution in different sparsity promoting norms. These methods conglomerate around two pivotal approaches; Low Rank Representation (LRR) \cite{liu2013robust} and Sparse Subspace Clustering (SSC) \cite{elhamifar2013sparse} methods benefiting from various imposed penalty conditions with respect to problem setup. The subspace clustering algorithm we utilized for the training dictionary matrix with a form of block Hankel structure is the Ordered Subspace Clustering (OSC) method \citealp{tierney2014subspace} which is derived from the SSC approach.

In SSC method, it is assumed that for a given observation matrix $X$, any column of which can be represented by the linear combination of the other similar columns. The objective function of this self similarity premise is given in the following equation where $Z$ is the coefficient matrix and $E$ is the error.

\begin{equation}
\label{eq:e8}
\min_{Z}\quad  \Arrowvert Z\Arrowvert_{1} \quad s.t. \quad X= XZ+E, \quad diag(Z)=0
\end{equation}

The sparsity promoting norm $\ell_{1}$ is used in the objective function.  After obtaining the coefficient matrix $Z$, an affinity matrix is composed and the clusters are labeled using spectral methods such as the Normalized Cuts \cite{shi2000normalized}. The OSC  incurs an additional penalty function in Eq. (\ref{eq:e8}) that enforces the similarity of the neighboring columns. In our problem setup, as the training matrix consists of neighboring columns with a block Hankel structure, we employ OSC in the training phase. The objective function of the OSC is given in Eq. (\ref{eq:e9}). 

\begin{multline}
\label{eq:e9}
\min_{Z, E}\quad \frac{1}{2} \Arrowvert E\Arrowvert_{F}^2+\lambda_{1}\Arrowvert Z\Arrowvert_{1} +\lambda_{2}\Arrowvert ZR\Arrowvert_{1,2}\\
\quad s.t. \quad X= XZ+E  
\end{multline}

where $R$ in the equation is a lower bidiagonal matrix, all the elements in the main and lower diagonal of which are respectively -1 and 1 as given below.  

\begin{equation} 
\label{eq:e10}
R =
\begin{bmatrix}
-1 &&&&&  \\
1 & -1 &&&&  \\
& 1&-1  &&& \\  
&& \ddots&\ddots&& \\  
&&&1&-1
\end{bmatrix}
\end{equation}

\begin{figure}[h]
 	\centering
 	\includegraphics[width=\columnwidth, keepaspectratio=true]{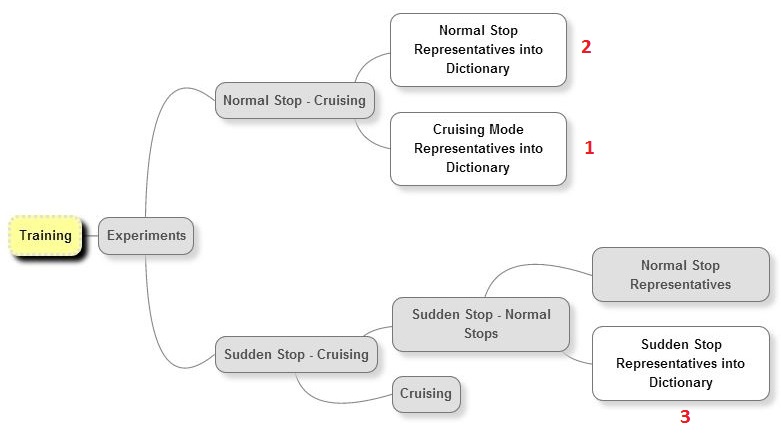}
 	\caption{Training Procedure}
 	\label{fig:trainscheme}	
 	\vspace{-1em}
 \end{figure}

The overall training scheme is given in Fig. (\ref{fig:trainscheme}). After collecting experimental data for each breaking states seven times, we build the training matrix using the data of only one experiment, the others are separated for test and simulations. It is not possible to conduct experiments for only one state, such as sudden stop, thus every experiment contains at least two breaking states such cruising and normal breaking.  

As shown in Fig. (\ref{fig:trainscheme}) in which the label number of the clusters are shown, we take an experiment in which cruising and normal breaking maneuver are subsequently repeated and cluster the obtained data only once assuming that there is no sudden breaking intention of the driver. However, we run the clustering algorithm twice for the experiment in which sudden stops are performed repeatedly to further eliminate the normal stops from the sudden stop cluster and obtain pure representatives for this state.  
 
\section{Implementation and Results}

\begin{figure}[h]
 	\centering
 	\includegraphics[width=\columnwidth, keepaspectratio=true]{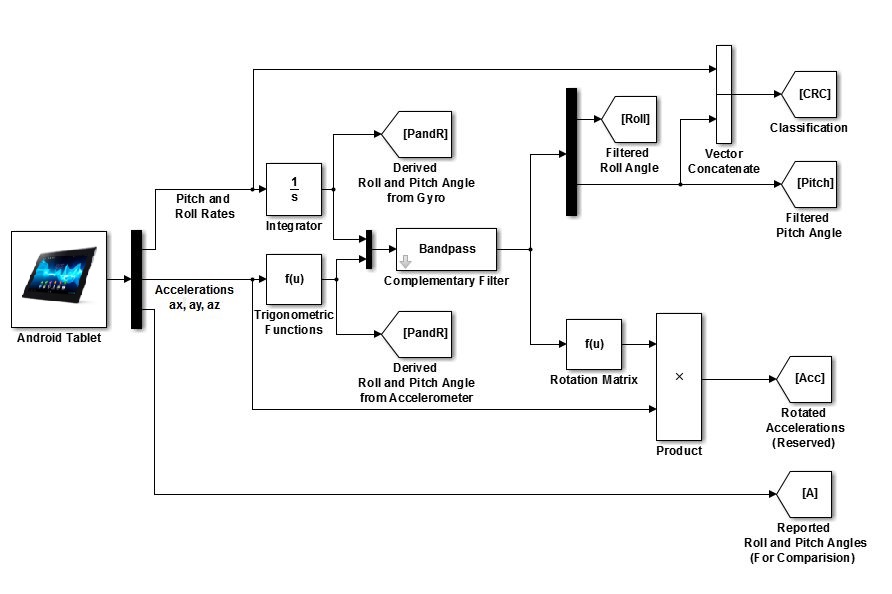}
 	\caption{Block Diagram of the Ride Assistance System - Signal Processing and Classification}
 	\label{fig:flow}
 	\vspace{-1em}	
\end{figure}

The procedures are summarized in Fig. (\ref{fig:flow}) prepared by Simulink toolbox in Matlab. We implemented all the procedures in Matlab for simulations and used Java programming environment for the mobile application. As the CRC\_RLS method has only a few lines of codes, the implementation does not take long for the mobile application once all the parameters and coefficients are readily available to the system. 
   
We give the results for remaining six experiments for each maneuver in the figure below. The normal and sudden breaking figures are put in the first and second column respectively. One of the dictionary features, the pitch angle of the handle bar is also seen in the figure to provide additional information. As seen in the first column, normal breaking results, the class of the maneuver is labeled at a frequency of 20 Hz and each experiments is a mean length of 1300 samples. In total the application labels the observed signal pattern 7800 times. The only misclassification region is seen in the first column on the fifth experiment which is enclosed by a circle. The breaking pattern of the rider in the enclosed region is close to sudden breaking where the pitch angle pattern is steeper then the others. The number of sudden breaking label on the figure is only six out of 7800 computations. On the sudden breaking column as we stated in the preceding sections, the results contain normal breaking labels.  
 
 \begin{figure}[h]
 	\centering
 	\includegraphics[width=\columnwidth, keepaspectratio=true]{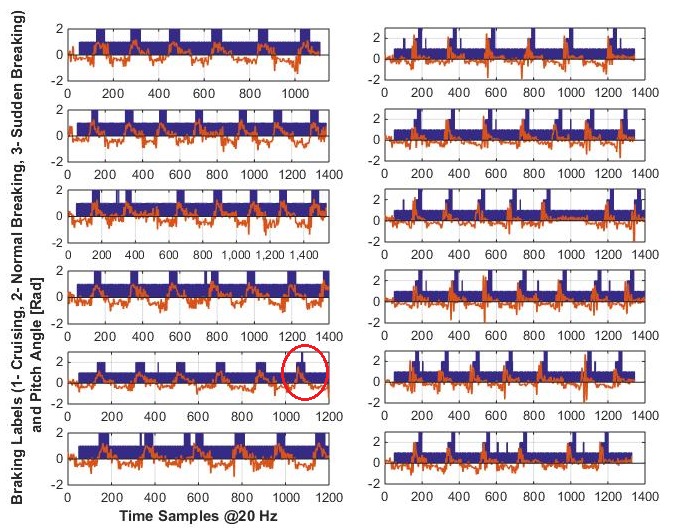}
 	\caption{Normal and Sudden Breaking Experiments Results}
 	\label{fig:results}
 	\vspace{-1em}	
 \end{figure}

\begin{figure}[h]
 	\centering
 	\includegraphics[width=\columnwidth, keepaspectratio=true]{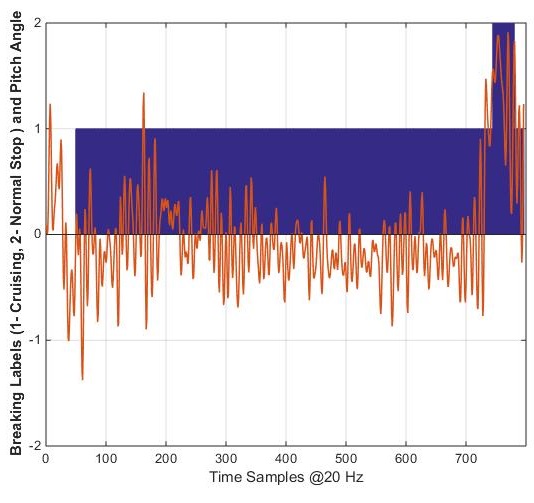}
 	\caption{Cruising Experiment Results}
 	\label{fig:cruising}
 	\vspace{-1em}	
 \end{figure}

 We also collected data for only cruising mode, the results of which is given in Fig. (\ref{fig:cruising}). In the 40 seconds experiment, the rider accelerates and reaches a constant velocity and stops at the end with the normal breaking pattern as seen in the figure. 
 
 We verified the accuracy and versatility of the proposed framework by following the same procedures for a different application. The authors in the study \cite{boyali2014block} compare B\_SRC and SRC methods and demonstrate steering a power wheel-chair by posture recognition \cite{boyali2014hand}. A hand continuously tracked by a Leap Motion sensor which accurately track the position and orientation of the hand at sub-millimeter levels and the recognized hand postures are mapped to the steering commands for a robotic wheelchair. We extended these studies by replacing the B\_SRC method for hand posture recognition with the CRC, employed the subspace clustering method in the training phase and introduce hand gesture recognition as the hand makes a gesture while in the transition state; changing position from one posture to another. We followed the same procedures given in this paper for the power wheelchair steering by hand posture and gesture recognition project in which we have been developing smart human machine interfaces for severely disabled people. 

\begin{figure}[h]
  	\centering
  	\includegraphics[width=\columnwidth, keepaspectratio=true]{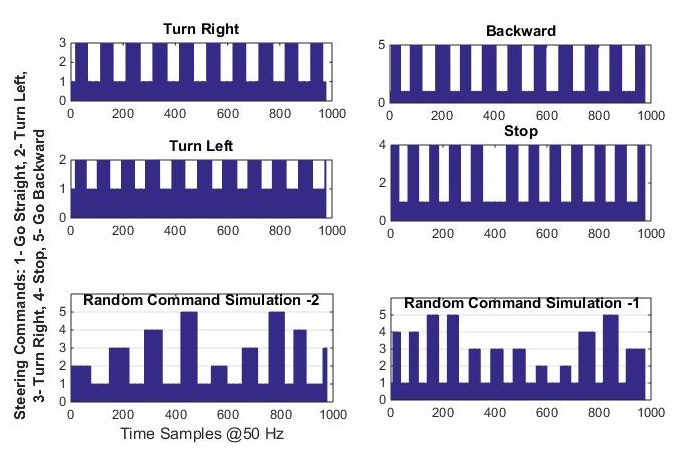}
  	\caption{Continous Hand Posture an Gesture Recognition for Steering a Power Wheelchair}
  	\label{fig:leap}
  	\vspace{-1em}	
\end{figure}  

The results of the experiments are given in Fig. (\ref{fig:leap}) which shows the recognition results of six different test simulations. In the simulations, a hand continuously change postures. There are five deictic hand posture states which are turn left, right, go straight, stop and go backward. The trajectory of the hand while it is changing the posture one to another is a gesture, the number of which is eight in the application. The application first classify whether the hand is in a gesture and posture state then label the hand state out of five postures and eight gestures.  The number of total classification is 5850 and the recognition accuracy is 100\% as shown in the figure. Unlike the braking state classification, the states of hand are well-defined and less noisy in the hand posture and gesture study in which the classification more accurate.

\section{Conclusion and Future Work}

In this study, we explained a framework to develop a ride assistance system which include a signal processing an classification module capable of classifying any type of signal patterns and in particular, we detailed the breaking mode classification and training procedures. The subspace method we employed in the training phases lead obtaining a clean dictionary in which the pure representative samples from each of the class are stacked a column vector.  Therefore the recognition accuracy is almost 100\% for the breaking modes and 100\% for the hand gesture and posture study in the real-time applications. In the near future, we plan to investigate the differences of the ride characteristics in terms of age, gender and skills with the robotic mobility and other mobility devices. 

\section*{Acknowledgment}
  
This study is supported by Japan Society for the Promotion of Science fellowship program. 
  
\nocite{*}
\bibliographystyle{IEEEtran} 
\bibliography{IEEEfull,mybibfile}

\end{document}